# Attention-Based Surface Representation Learning for Robot State Prediction and Open-Ended Surface Classification*

Oleg Kushnarev, Alexander Belyaev

**Abstract—For ground robots operating in outdoor environments, understanding the properties of the underlying terrain is essential for ensuring reliable operation. In most perception-based studies, this problem is formulated as categorical classification with a fixed number of classes defined during training. We propose an approach that enables new surface classes to be added as trainable vectors, which can subsequently be used to address higher-level tasks. By employing a learning paradigm based on predicting the robot's next state in time and using attention blocks, we improved classification accuracy to 98.56% on the Belyaev–Kushnarev dataset and 94.8% on BorealTC.**

## I. Introduction

The operational range of a mobile robot is fundamentally limited by its interaction with the supporting surface. In real-world conditions, the robot encounters a continuous spectrum of deformable, loose, and rigid surfaces. Changes in ground surface (asphalt → gravel → wet clay → ice) drastically alter the mechanics of wheel-ground contact, manifesting as slippage, increased rolling resistance, and loss of traction [1]–[4]. As a result, the robot is constantly at risk of getting stuck, drive saturation, and mission failure. The ability to characterize the supporting surface in advance becomes a prerequisite for reliable autonomy in unstructured scenarios.

To address this challenge, two sensor-based approaches have historically emerged: exteroception and proprioception [5], [6]. Exteroceptive solutions (cameras [7], [8], lidars [6], radars [9], and infrared sensors [10]) provide a rich semantic context, allowing the robot to identify "grass" or "sand" based on visual texture [11]–[13]. However, they capture the surface's appearance rather than its dynamic characteristics. Visual cues are subject to variability, wet asphalt and dry ice may look similar, whereas dry sand and loose gravel are visually different but dynamically equivalent. Moreover, deteriorating environmental conditions (dust, fog, precipitation) render exteroceptive sensors unreliable precisely when mobility is most at risk, whereas proprioceptive signals remain stable [14].

In contrast, proprioceptive perception, data from encoders, motor currents, and IMUs, provides a direct physical assessment of interaction with the ground [5], [15], [16]. This channel reflects the cause-and-effect relationship between a command and movement. Existing proprioceptive methods can be divided into several groups: vibrational (IMU/accelerometers) [17]–[21]; contact-based (strain gauges on wheels or legs) [22]–[26]; and energy-based (motor currents/torques) [27]–[30]. However, regardless of the group, most approaches fall firmly into two paradigms: (i) categorical classification (grass, asphalt, ice) [17], [29], [31], [32], based on the unrealistic assumption that properties are static within a class; and (ii) regression of state variables (slip, traversability) [1], [2], [33]. Both paradigms are critically dependent on extensive labeled reference data, which is extremely difficult to obtain under real-world conditions.

To overcome these limitations, we propose an alternative approach to proprioceptive perception. Instead of focusing solely on classification or regression, we formulate the problem as self-supervised learning based on time series: the model is trained to predict the next state of the sensors (currents, angular velocities, IMU) based on a window of historical data, and the prediction error serves as a signal for self-supervised learning. Since the true next state is always available in the data stream, training does not require external labels. In addition, we introduce a vector into the architecture that describes the surface model, which is used for prediction. This vector will be stored in the robot's memory, thereby incorporating models responsible for different classes. Focusing on the concept of model diversity allows us to describe the distinct surfaces the robot encounters during operation. In other words, it enables us to simultaneously solve a classification problem with an expandable number of classes using trainable model vectors. The parameters of the model vectors are integrated into the dynamic model: the same control input leads to different sensory responses on different surfaces, and the model is forced to adjust its parameters to minimize prediction error. Thus, the parameters implicitly encode the variability of surface properties. To our knowledge, this is the first work to combine self-learning based on temporal prediction with trainable parametric models for proprioceptive surface classification. Unlike most existing approaches, we focus on energy-based classification methods [27]–[29], which remain the least studied despite the widespread availability of current sensors on autonomous robots [33]–[36].

Validation was performed both on a publicly available datasets [27], [29]. The results show that the learned representation effectively classifies soil types under zero-transfer conditions and can be used in a predictive controller to prevent getting stuck.

## II. Methods

As part of our solution, we sought to address the following issues in classifying surface types: reliance solely on loss

*The research was funded by the Russian Science Foundation, No. 26-79- 00101, https://rscf.ru/project/26-79-00101/.

Equal contribution. All authors are with the National Research Tomsk Polytechnic University, Tomsk, 634050 Russia.

Corresponding author: Belyaev A.

Emails: {okushnarev, asb22}@tpu.ru

functions for supervised classification, a fixed number of classes, and a lack of understanding of surface properties.

In our solution, we use an encoder-decoder architecture based on attention blocks with some modifications that allow us to identify and extract vectors describing surface models. The model's architecture is shown in Fig. 1.

### A. Encoder

We feed tokens describing the robot's states $\mathbf{x}$ over a time interval of length H starting from the current time $t$, [$\mathbf{x}^{t-H}$ , ... , $\mathbf{x}^{t}$ ], into the encoder's input. We determine the robot's states based on the sensors available on the robot. Additionally, we feed the robot's action values $\mathbf{a}$ for the same interval [$\mathbf{a}^{t-H}$ , ... , $\mathbf{a}^{t}$ ]; this is necessary so that the model can understand the changing input that affects the value of the vector $\mathbf{x}^{t} - \mathbf{x}^{t-H}$ **.** The final component fed into the encoder is a specialized token called <*surf*>. This token is trainable and is used to describe the surface based on sensor readings. All these vectors pass through preliminary MLP layers and are fed into the Self-attention block. This block is primarily designed to enrich the <*surf*> vector with important information from the sensors. This results in a vector that describes the surface model along which the robot moved during the given time interval H.

### B. Decoder

The transformed information, in the form of vectors $\mathbf{z}$ and the updated vector <*surf*>, is fed to the decoder (input Q), which consists of cross-attention and MLP layers. In parallel, a special trainable token sequence $q$ is passed through self-attention and then fed into inputs K and V. For now, we are feeding only a single token $q$; however, by analogy with Transformer architectures, the number of tokens can be increased to enable sequential prediction several steps ahead. At the decoder's output, after passing through the MLP, a vector $\mathbf{x}^{t+1}$ is formed, describing the robot's state at the next time step**.** The loss for this part of the network is defined as the MSE metric (1).

$$\mathcal{L}_{\text{MSE}} = \frac{1}{n}\sum_{i=1}^{n}\left(\mathbf{x}^{t+1} - \hat{\mathbf{x}}^{t+1}\right)^2 \tag{1}$$

### C. Surface Models

In parallel with the decoder, we use the updated <*surf*> vector to generate a library of surface models ($s_n$). We feed it into a dedicated memory where we already store previously obtained surface models {$s_1$ , ..., $s_n$}. This vector is compared with the vectors of existing surfaces using pairwise cosine similarity. An additional loss based on the metrics of known surface classes is calculated using cross-entropy. We drew inspiration for the implementation and training of this block from the ideas of Prototypical Contrastive Learning [37], but deviated from self-supervised learning by using labeled classes for training (2).

$$\mathcal{L}_{\text{proto}} = -\log\left(\frac{\exp\left(\text{sim}\left(surf, s_n\right)/\tau\right)}{\sum_{i=1}^{n}\exp\left(\text{sim}\left(surf, s_i\right)/\tau\right)}\right), \tag{2}$$

where *sim(...)* is cosine similarity between two vectors, $\tau$ is the temperature coefficient that controls sharpness the similarity-based distribution over classes.

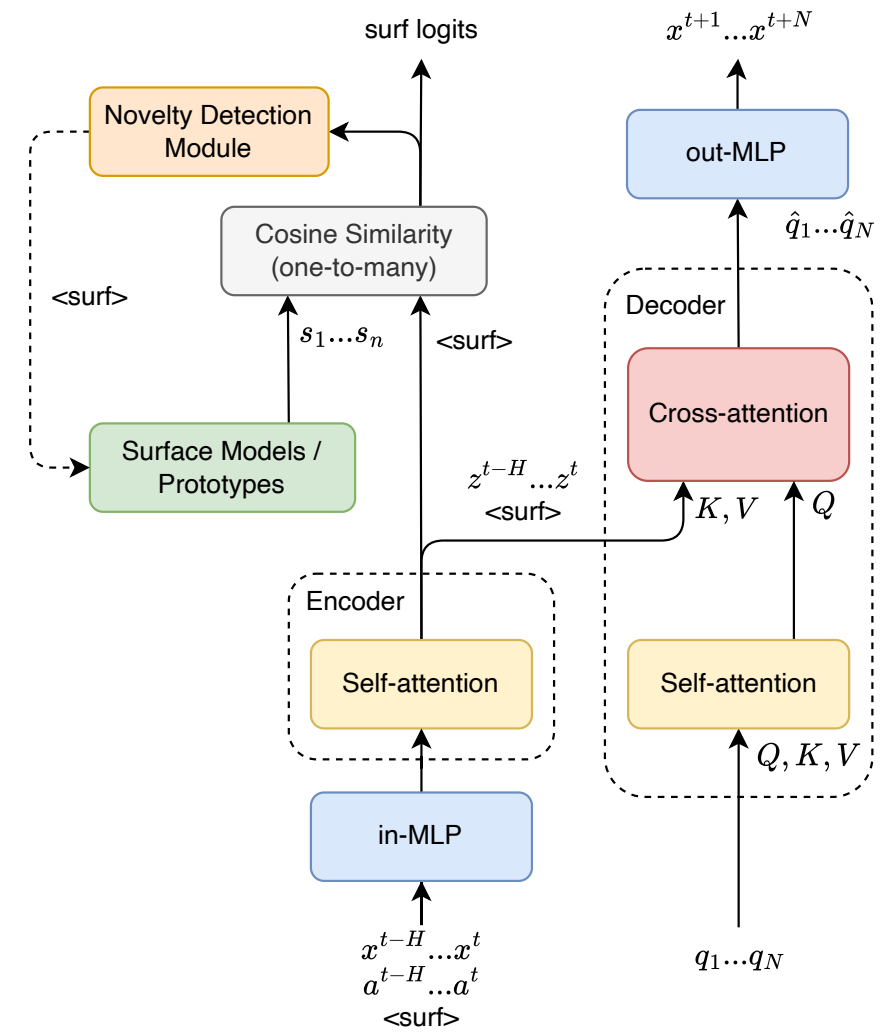

Figure 1. Architecture of the proposed model. The encoder processes a temporal sequence of robot states $\mathbf{x}$, actions $\mathbf{a}$, and a trainable <*surf*> token using MLP and self-attention layers to obtain latent representations $\mathbf{z}$ and a surface representation. The decoder combines these representations with a trainable query token $\mathbf{q}$ through self- and cross-attention layers and MLPs to predict the robot state $\mathbf{x}^{t+1}$.

As a result, the network's total loss consists of two parts and is expressed as in

$$\mathcal{L} = \lambda_{\text{MSE}} \cdot \mathcal{L}_{\text{MSE}} + \lambda_{\text{proto}} \cdot \mathcal{L}_{\text{proto}}, \tag{3}$$

where $\lambda_{\text{MSE}}$, $\lambda_{\text{proto}}$ are loss weigths.

### D. Descriptions of the learning process

For all datasets, model development was organized into three successive phases: dataset preparation, hyperparameter optimization, and final model training. The first phase involved preprocessing the data and defining the corresponding data splits. This was followed by an automated search for a suitable set of hyperparameters, after which the model was trained using the configuration selected during optimization.

To prepare the data, each original dataset was initially divided into relatively large, mutually exclusive segments. This segmentation was performed in a way that preserved the temporal or sequential order of observations within the individual datasets. The resulting segments were subsequently allocated to the training, validation, and test sets in a 70:10:20 ratio, while preserving the distribution of the target classes. Segment dimensions were adapted to the characteristics of each dataset, particularly its overall size and class balance. Once the splits had been established, model inputs were constructed independently within each segment using a rolling-window procedure.

Hyperparameter optimization was performed using Optuna. For each dataset, 100 trials were evaluated, with a maximum budget of 30 training epochs per trial. To reduce computational overhead, the HyperBand pruning strategy was applied to terminate trials whose performance indicated limited potential for improvement.

Following optimization, the network was trained for a maximum of 1,200 epochs. Model checkpoints were evaluated on the validation set at intervals of 5 epochs, and the best-performing checkpoint was retained. Training used the AdamW optimizer, while the learning rate was dynamically adjusted using a ReduceLROnPlateau scheduler.

TABLE I. CLASSIFICATION REPORT ON THE BOREALTC DATASET

| Surface | Precision (%) | Recall (%) | F1-score (%) | Accuracy (%) |
|---|---|---|---|---|
| asphalt | 94.13±0.88 | 82.65±2.01 | 88.01±1.04 | 94.80 ± 0.15 |
| flooring | 97.23±0.99 | 96.55±0.83 | 96.88±0.10 | |
| ice | 96.30±0.35 | 97.79±0.84 | 97.03±0.26 | |
| silty loam | 89.28±0.93 | 90.96±1.90 | 90.09±0.60 | |
| snow | 91.67±1.99 | 93.91±0.84 | 92.76±0.75 | |

## III. RESULTS

To test the algorithm, we used two datasets based on perception data: BorealTC [27] and Belyaev-Kushnarev dataset [29]. To ensure reliable statistics, the same model was trained on each dataset using three different random seeds.

### A. Results on the BorealTC dataset

The paper [27] presents data collected in outdoor conditions, specifically in boreal forests. The authors used a Husky A200 four-wheeled robot capable of traveling on five different types of terrain (asphalt, flooring, ice, silty loam, and snow). For this dataset, the state vector **x** included data from two motor current sensors ($I_L$, $I_R$), the rotational speed of the two wheels as determined by optical encoders ($\omega_L$, $\omega_R$), an IMU (3 angular velocities $\{\omega_x, \omega_y, \omega_z\}$ and 3 accelerations $\{a_x, a_y, a_z\}$), and the calculated relative energy expenditure coefficient (*Ke*). Unfortunately, this dataset lacks information on the set values of motor rotational speed or angular or linear motion; therefore, the action vector **a** will be empty.

The quality metrics of the trained model across three independent training runs are presented in Table I. The model's classification accuracy is 94.8%, which is higher than the 93.96% reported by the authors in their original paper [27].

Metrics for prediction of the robot's next state in time are shown in Table II. Notably, while classification metrics indicated that the model began to overfit relatively early in training, regression metrics suggested that training continued throughout the entire training period. This discrepancy implies that there may be scope for further improving accuracy in this domain.

TABLE II. REGRESSION REPORT ON THE BOREALTC DATASET

| State | MAE | RMSE | R² |
|---|---|---|---|
| $\omega_x^{t+1}$ | 0.0097 ± 0.0003 | 0.0176 ± 0.0008 | 0.8076 ± 0.0166 |
| $\omega_y^{t+1}$ | 0.0095 ± 0.0004 | 0.0161 ± 0.0005 | 0.8746 ± 0.0082 |
| $\omega_z^{t+1}$ | 0.0240 ± 0.0021 | 0.0366 ± 0.0026 | 0.9890 ± 0.0015 |
| $a_x^{t+1}$ | 0.7025 ± 0.0145 | 1.2870 ± 0.0173 | 0.5468 ± 0.0121 |
| $a_y^{t+1}$ | 0.7838 ± 0.0133 | 1.4012 ± 0.0120 | 0.4703 ± 0.0090 |
| $a_z^{t+1}$ | 0.3906 ± 0.013 | 0.7046 ± 0.0119 | 0.2942 ± 0.0236 |
| $\omega_L^{t+1}$ | 0.0239 ± 0.0026 | 0.0644 ± 0.0084 | 0.9559 ± 0.0113 |
| $\omega_R^{t+1}$ | 0.028 ± 0.0039 | 0.0997 ± 0.0098 | 0.8988 ± 0.0200 |
| $I_L^{t+1}$ | 0.8994 ± 0.0452 | 1.6021 ± 0.0449 | 0.8446 ± 0.0087 |
| $I_R^{t+1}$ | 0.8834 ± 0.1017 | 2.9303 ± 0.3176 | 0.6523 ± 0.0724 |
| $Ke^{t+1}$ | 0.53.71 ± 0.0467 | 2.1155 ± 0.0043 | 0.5968 ± 0.0016 |

TABLE III. CLASSIFICATION REPORT ON BELYAEV-KUSHNAREV DATASET

| Surface | Precision (%) | Recall (%) | F1-score (%) | Accuracy (%) |
|---|---|---|---|---|
| brown | 100 | 100 | 100 | 98.56 ± 0.20 |
| gray | 97.20 ± 0.44 | 98.80 ± 1.05 | 97.99 ± 0.30 | |
| green | 97.24 ± 1.03 | 97.18 ± 0.48 | 98.35 ± 0.26 | |
| table | 100 | 100 | 100 | |

### B. Results on Belyaev-Kushnarev dataset

The paper [29] presents experiments conducted in indoor conditions that simulate the effects of outdoor surfaces. The experiments use a robot with three omni-directional wheels moving across four types of surfaces: soft, smooth rubber (encoded as gray); hard, rough plastic (green); hard, smooth laminated chipboard (table); and medium-stiffness, rough, honeycomb-shaped EVA (brown). For this dataset, the state vector **x** includes data from three motor current sensors $\{I_1, I_2, I_3\}$, the rotational speeds of the three wheels as determined by optical encoders ($\omega_1$, $\omega_2$, $\omega_3$), and the relative energy consumption coefficient (*Ke*). The action vector **a** is formed by the specified motor rotational speeds ($\omega_1^{set}$, $\omega_2^{set}$, $\omega_3^{set}$).

The quality metrics of the trained model across three independent training runs are presented in Table III. The classification accuracy, as measured by the accuracy metric, is 98.56%, which exceeds the 91% accuracy reported in the original paper.

At the same time, the accuracy of predicting the robot's next state, as measured by both the MAE and $R^2$ metrics, is high, as shown in Table IV.

The values of the surface vectors themselves are of particular interest from an analytical standpoint. To analyze them, we used the t-SNE decomposition shown in Fig. 2 and Fig. 3. On Belyaev-Kushnarev dataset, the generated surface vectors exhibit nearly perfect separability in space, which further confirms the high accuracy of classification and the separation of surface properties in feature space. On the BorealTC dataset, the three main surface types, snow, ice, and flooring, differ significantly in their features. However, surfaces such as asphalt and sandy loam are located in transition zones between other surfaces.

TABLE IV. REGRESSION REPORT ON BELYAEV-KUSHNAREV DATASET

| State | MAE | RMSE | R² |
|---|---|---|---|
| $\omega_1^{t+1}$ | 3.5676 ± 0.4622 | 4.7775 ± 0.5953 | 0.9848 ± 0.0038 |
| $\omega_2^{t+1}$ | 3.1819 ± 0.4060 | 4.3899 ± 0.5686 | 0.9859 ± 0.0036 |
| $\omega_3^{t+1}$ | 3.3783 ± 0.5131 | 4.5401 ± 0.5947 | 0.9863 ± 0.0037 |
| $I_1^{t+1}$ | 0.1212 ± 0.0017 | 0.1546 ± 0.0012 | 0.8541 ± 0.0023 |
| $I_2^{t+1}$ | 0.1129 ± 0.0006 | 0.1444 ± 0.0012 | 0.8611 ± 0.0023 |
| $I_3^{t+1}$ | 0.1182 ± 0.0016 | 0.1521 ± 0.0009 | 0.8577 ± 0.0016 |
| $K_e^{t+1}$ | 0.5485 ± 0.0010 | 0.7019 ± 0.0058 | 0.8476 ± 0.0025 |

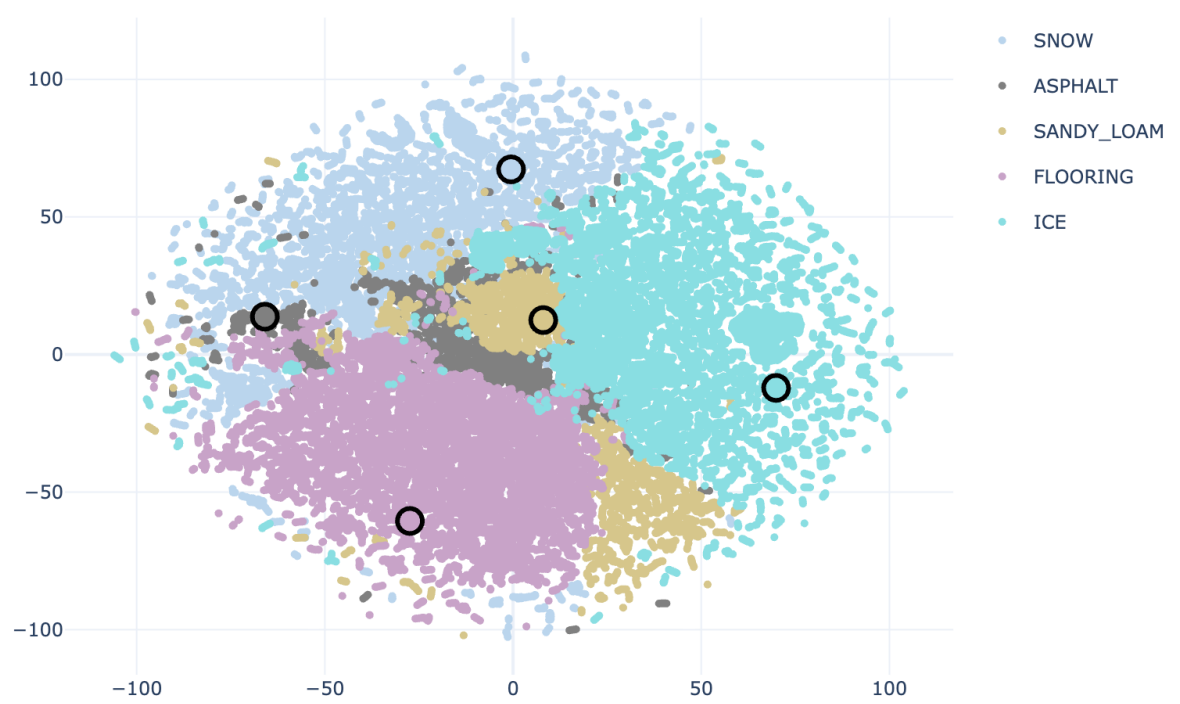

Figure 2. An illustration of class proximity using t-SNE analysis on the borealtc dataset. Each inset illustrates a terrain class in the dataset. The black circles denote the vectors that, during the training process, formed the vectors for each of the surfaces and were stored in memory.

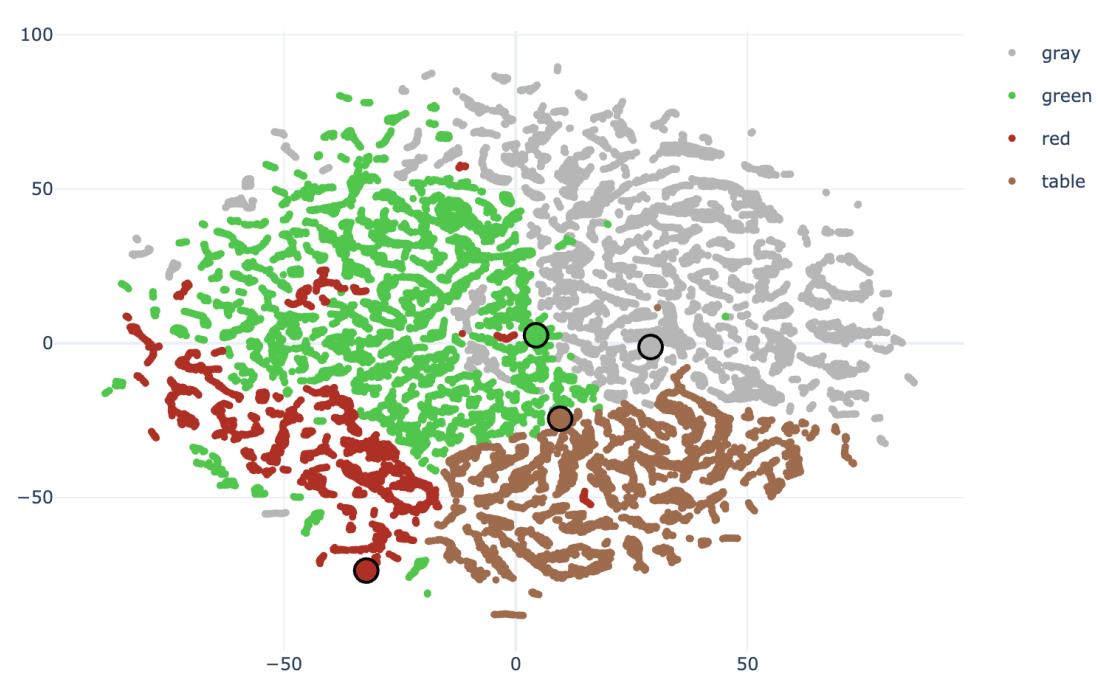

Figure 3. T-SNE decomposition of the surface vectors obtained on Belyaev-Kushnarev dataset. Black circles denote the vectors that, during training, formed the vectors for each of the surfaces and were stored in memory.

Furthermore, sandy loam is found in several regions at once, indicating the heterogeneity of its characteristics and the complexity of the surfaces, which is confirmed by its physical description in the original study. From all of this, we conclude that the surface vectors obtained using this method capture the physics of their properties. However, this requires further, more in-depth research.

## IV. Conclusion

In this paper, we propose a solution to the problem of classifying ground surfaces for outdoor mobile robots, transitioning from a categorical classification approach, limited by the number of classes, to an open-ended model-based approach. We were able to improve the performance of perception-based approaches on two datasets. In the future, we plan to expand this approach as part of the transition to fully self-supervised learning. Additionally, we would like to analyze the resulting surface models in greater depth, with the goal of interpreting them for further application in more complex mobile robotics tasks, such as local navigation or safe route planning. Project is available at https://github.com/okushnarev/transformer-cls